\documentclass[10pt,twocolumn,letterpaper]{article}

\usepackage{cvpr}              

\newcommand{\TODO}[1]{\textbf{\color{red}[TODO: #1]}}
\renewcommand{\TODO}[1]{}

\usepackage{microtype}

\definecolor{cvprblue}{rgb}{0.21,0.49,0.74}
\usepackage[pagebackref,breaklinks,colorlinks,allcolors=cvprblue]{hyperref}

\usepackage{minted}
\usepackage{multirow}
\usepackage{arydshln}
\usepackage{algorithm}
\usepackage{algorithmicx}
\usepackage{algpseudocode}
\usepackage{amssymb}
\usepackage{booktabs}
\usepackage{multirow}
\usepackage{colortbl}
\usepackage{pifont}
\usepackage{graphicx}
\usepackage{amsmath}
\usepackage{enumitem}
\usepackage[most]{tcolorbox}

\usepackage{marvosym}

\usepackage[table]{xcolor} 

\usepackage{paracol}
\usepackage{tcolorbox}
\tcbset{enhanced, sharp corners}
\usepackage{tikz}
\usepackage[utf8]{inputenc}
\usepackage{graphicx}
\usepackage[dvipsnames]{xcolor}
\usepackage{tcolorbox}
\tcbuselibrary{skins, raster}

\usepackage{array}
\usepackage{placeins}
\usetikzlibrary{positioning,arrows.meta,shapes,fit}

\def\paperID{37810} 
\def\confName{CVPR}
\def\confYear{2026}

\title{EchoTrail-GUI: Building Actionable Memory for GUI Agents via Critic-Guided Self-Exploration}

\author{
    Runze Li$^{1,2}$\thanks{Equal contribution.} \thanks{Work was done during an internship at Alibaba group.} \quad
    Yuwen Zhai$^{1}$\footnotemark[1]\, \textsuperscript{\Letter} \quad
    Bo Xu$^{1}$ \quad
    Liwu Xu$^{1}$ \quad
    Nian Shi$^{1}$ \\
    Wei Zhang$^{2,3}$ \quad
    Ran Lin$^{1}$ \quad
    Liang Wang$^{1}$\textsuperscript{\Letter}
    \\[0.8mm]
    \small
    $^{1}$Taobao \& Tmall Group of Alibaba \quad
    $^{2}$East China Normal University \quad
    $^{3}$Shanghai Innovation Institute \\[0.3mm]
    \scriptsize
    51265901021@stu.ecnu.edu.cn, zhangwei.thu2011@gmail.com, \\
    \scriptsize\{zhaiyuwen.zyw, songbai.xb, xuliwu.xlw, shinian.sn, linran.lr09, liangwang.wl\}@alibaba-inc.com
}

\begin{document}
\maketitle

\begingroup
\renewcommand{\thefootnote}{}
\footnotetext{\textsuperscript{\Letter}Corresponding author.}
\endgroup

\begin{abstract}
Contemporary GUI agents, while increasingly capable due to advances in Large Vision-Language Models (VLMs), often operate with a critical limitation: they treat each task in isolation, lacking a mechanism to systematically learn from past successes. This ``digital amnesia'' results in sub-optimal performance, repeated errors, and poor generalization to novel challenges. To bridge this gap, we introduce EchoTrail-GUI, a novel framework designed to mimic human-like experiential learning by equipping agents with a dynamic, accessible memory. Our framework operates in three distinct stages. First, during Experience Exploration, an agent autonomously interacts with GUI environments to build a curated database of successful task trajectories, validated by a reward model. Crucially, the entire knowledge base construction is thus fully automated, requiring no human supervision. Second, in the Memory Injection stage, upon receiving a new task, our system efficiently retrieves the most relevant past trajectories to serve as actionable ``memories.'' Finally, during GUI Task Inference, these memories are injected as in-context guidance to inform the agent's reasoning and decision-making process. We demonstrate the efficacy of our approach on benchmarks including Android World and AndroidLab. The results show that EchoTrail-GUI significantly improves the task success rate and operational efficiency of baseline agents, validating the power of structured memory in creating more robust and intelligent GUI automation.
\end{abstract}  
\section{Introduction}
\label{sec:intro}

The rapid advancement of Large Vision-Language Models (VLMs)~\cite{bai2025qwen2, wang2024cogvlm} has fundamentally reshaped the development of autonomous GUI agents. Early approaches relied on structured metadata such as accessibility trees~\cite{wen2024autodroid}, whereas modern systems can now parse raw GUI screenshots and follow complex user instructions through sequences of fine-grained actions, including clicking, scrolling, typing, and navigation~\cite{hong2024cogagent, qin2025ui, xu2024aguvis}. These capabilities have enabled significant progress in digital task automation, intelligent computer-use assistants, and interactive multimodal copilots.

Despite these advances, existing GUI agents still suffer from a core limitation that we term \emph{digital amnesia}. Most current agents remain stateless: they approach each task independently, lacking any mechanism to accumulate operational knowledge or capitalize on prior successful behaviors. This deficiency restricts their robustness, leads to repeated mistakes, and undermines generalization when encountering diverse application layouts or multi-step tasks. Two fundamental obstacles contribute to this limitation.

The first is the \emph{experience acquisition bottleneck}. High-quality trajectory data is crucial for building generalizable GUI agents~\cite{rawles2024androidworld, xu2024androidlab}, yet manual annotation is unscalable and unguided exploration produces low-utility traces. Recent efforts synthesize trajectories from videos~\cite{jang2025scalable}, web tutorials~\cite{xu2024agenttrek, zhang2025tongui}, or reverse task inference~\cite{sun2025genesis}, while others explore autonomous exploration~\cite{xie2025gui} but typically lack effective quality control mechanisms.

The second is the \emph{knowledge application gap}. Even when trajectory corpora exist, most GUI agents rely on static examples or handcrafted prompts~\cite{yang2023set, yang2025aria}, failing to retrieve and apply past experience dynamically. Retrieval-augmented methods~\cite{xu2025retrieval} are promising but depend on well-curated, semantically aligned memories that are typically scarce.

\begin{figure*}[t]
    \centering
    \includegraphics[width=1.0\textwidth]{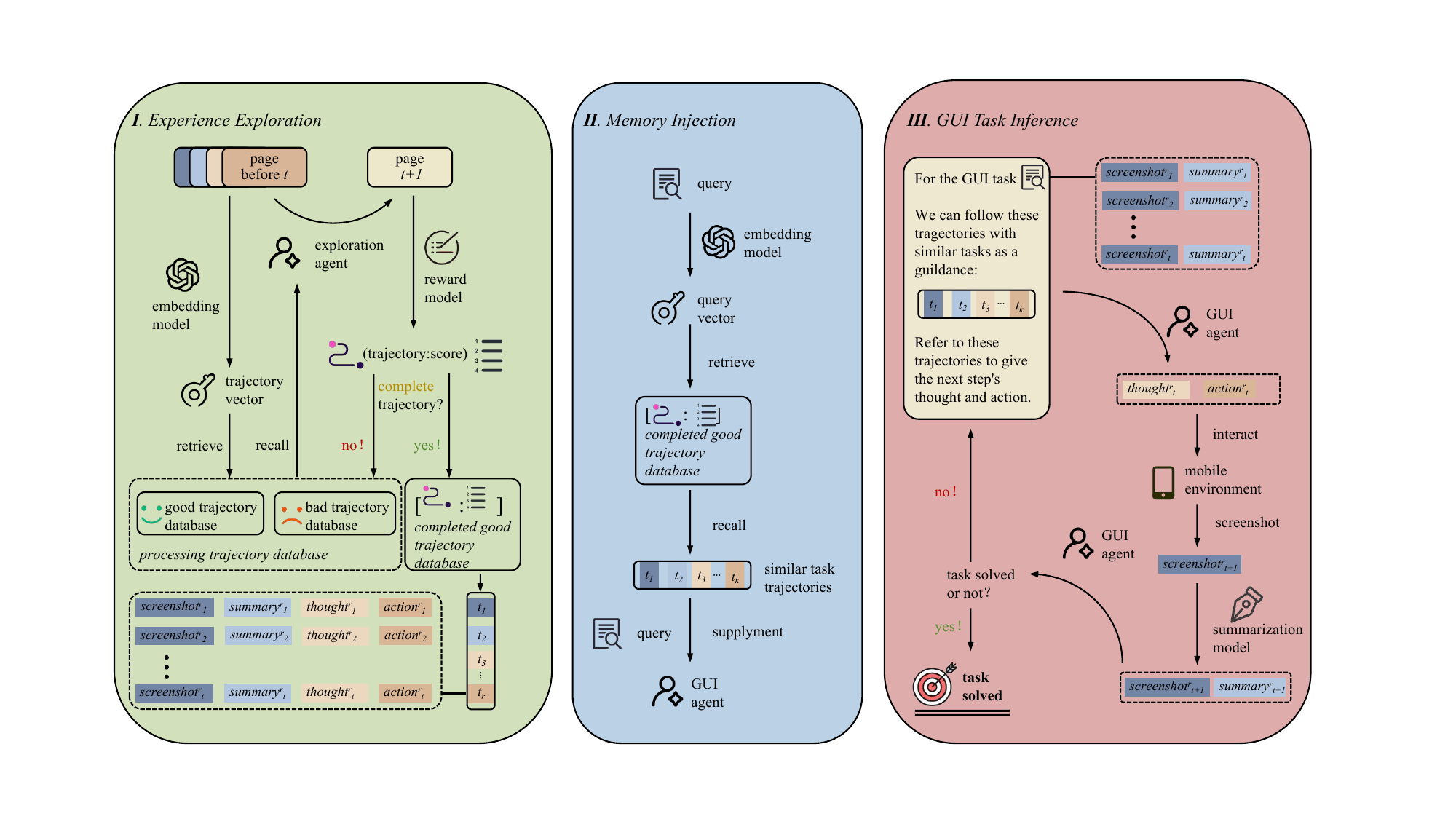}
    \caption{\textbf{The Architecture of EchoTrail-GUI.} Our framework consists of three stages. \textbf{(I) Critic-Guided Self-Exploration:} An exploration agent \(\pi_{\text{explore}}\) generates trajectories (\(\tau\)) which are evaluated by a critic (\(R_{\text{critic}}\)). In-progress trajectories guide exploration via a processing database (\(D_{\text{proc}}\)), while high-quality completed trajectories are archived into a permanent memory database (\(D_{\text{mem}}\)). \textbf{(II) Dynamic Memory Injection:} For a new task instruction \(I\), a hybrid retrieval system fetches the top-\(K\) relevant trajectories \(M_t\) from \(D_{\text{mem}}\) to serve as contextual memory. \textbf{(III) Memory-Augmented Inference:} The retrieved memories \(M_t\) are injected into the prompt of a GUI agent \(\pi_{\text{aug}}\), guiding its iterative process of thought, action, and observation to solve the task.}
    \label{fig:framework_overview}
\end{figure*}

To address these challenges, we introduce \textbf{EchoTrail-GUI}, a unified framework that emulates the human-like cognitive cycle of \emph{learn–remember–apply}. Our central goal is to construct a self-improving, closed-loop system where the agent autonomously gathers experience, distills actionable knowledge, and reuses it to guide new tasks. Inspired by retrieval-augmented generation (RAG)~\cite{lewis2020retrieval}, EchoTrail-GUI enables GUI agents to dynamically access the most relevant and successful past trajectories as structured, contextual guidance.

As illustrated in Figure~\ref{fig:framework_overview}, EchoTrail-GUI consists of three stages:

\begin{enumerate}
\item \textbf{Experience Exploration.} An exploration agent autonomously interacts with the GUI environment to generate diverse task trajectories. A critic-style reward model evaluates trajectory quality, allowing the system to construct a high-fidelity, human-free repository of successful trajectories.
\item \textbf{Memory Injection.} Given a new task instruction, we retrieve semantically relevant past experiences using a hybrid dense--sparse retrieval strategy. These trajectories serve as contextual memory that reflects the agent’s own operational history.
\item \textbf{GUI Task Inference.} Retrieved trajectories are reformatted into structured guidance and injected into the agent’s prompt, acting as an operational blueprint that supports reasoning, reduces redundant exploration, and improves action prediction.
\end{enumerate}

By integrating these stages into a fully automated pipeline, EchoTrail-GUI transforms stateless GUI agents into memory-augmented systems that grow more capable over time.

In summary, our contributions are as follows:

\begin{itemize}
\item We present EchoTrail-GUI, a three-stage framework that autonomously constructs and leverages actionable memory for GUI agents through critic-guided self-exploration, dynamic retrieval, and memory-augmented inference.
\item We build \textbf{EchoTrail-4K}, a curated dataset containing over 4,000 high-quality trajectories collected in Android environments, supporting both retrieval-based inference and downstream training.
\item Through extensive experiments on AndroidWorld~\cite{rawles2024androidworld} and AndroidLab~\cite{xu2024androidlab}, we demonstrate that memory-augmented GUI agents significantly outperform their stateless counterparts, yielding substantial gains in success rate, sub-goal completion, and operational robustness.
\end{itemize}

\section{Related Work}
\label{sec:relat}

GUI agents are designed to perceive and interact with digital environments by processing visual screen information. The emergence of powerful Multimodal Large Language Models (MLLMs)~\cite{achiam2023gpt, bai2025qwen2, anthropic2024Claude} has fundamentally shifted the paradigm from relying on structured metadata (like accessibility trees or XML) to end-to-end approaches that directly process raw screen pixels~\cite{rawles2024androidworld, wanyan2025look, lai2025androidgen}. Research in this area is primarily explored along two vectors. One major research thrust focuses on enhancing the intrinsic capabilities of a single powerful agent. This involves extensive pre-training on large, GUI-centric datasets~\cite{qin2025ui} or proposing unified action spaces to improve interaction abilities~\cite{wu2024atlas}. A parallel, training-free strategy leverages the in-context learning and reasoning of MLLMs, often combining reasoning approaches like ReAct~\cite{yao2022react} with visual annotation techniques~\cite{rawles2024androidworld, yang2023set}. The second vector explores the efficacy of collaborative multi-agent systems. These frameworks typically employ an orchestrator-agent architecture to decompose complex tasks and delegate sub-tasks~\cite{fourney2024magentic, lai2025androidgen}. In the GUI domain, this has been extended to integrating specialized modules such as reflection agents for performance evaluation and error correction~\cite{wang2024mobile}, or self-evolving modules that utilize web knowledge and episodic memory for adaptation~\cite{zhang2025appagent}.

\begin{algorithm}[t]
\caption{Critic-Guided Self-Exploration}
\label{alg:exploration}
\begin{algorithmic}[1]
\State \textbf{Input:} Exploration agent \(\pi_{\text{explore}}\), Critic \(R_{\text{critic}}\), Threshold \(\theta_{\text{good}}\)
\State \textbf{Initialize:} Processing DB \(D_{\text{proc}} \leftarrow \emptyset\), Memory DB \(D_{\text{mem}} \leftarrow \emptyset\)
\For{episode \(i = 1, \dots, N\)}
    \State \(\tau_{\text{new}} \leftarrow (s_0)\) where \(s_0\) is the initial state.
    \For{step \(t = 0, \dots, T_{\text{max}}-1\)}
        \State \(H_t \leftarrow\) actions in \(\tau_{\text{new}}\)
        \State \(G_t \leftarrow \text{RetrieveGuidance}(D_{\text{proc}}, \tau_{\text{new}})\) 
        \State \(a_t \leftarrow \pi_{\text{explore}}(s_t, H_t, G_t)\)
        \State Execute \(a_t\), observe next state \(s_{t+1}\)
        \State \(\tau_{\text{new}} \leftarrow \tau_{\text{new}} \circ (a_t, s_{t+1})\) \Comment{Append to trajectory}
        \State Update \(D_{\text{proc}}\) with the partial trajectory \(\tau_{\text{new}}\)
        \If{stopping condition is met} \textbf{break} \EndIf
    \EndFor
    \If{\(R_{\text{critic}}(\tau_{\text{new}}) \geq \theta_{\text{good}}\)}
        \State \(D_{\text{mem}} \leftarrow D_{\text{mem}} \cup \{\tau_{\text{new}}\}\) \Comment{Archive high-quality memory}
    \EndIf
\EndFor
\State \textbf{Return:} Memory Database \(D_{\text{mem}}\)
\end{algorithmic}
\end{algorithm}

The creation of high-fidelity trajectory data is a critical prerequisite for advancing GUI agents, as this data underpins their ability to plan and execute tasks in complex digital worlds~\cite{pan2024autonomous}. The challenges associated with manual annotation have accelerated research into automated data generation. One approach involves synthesizing trajectories from existing human-generated content. This includes extracting (frame, action) sequences from online videos~\cite{jang2025scalable} or converting web tutorials into structured, synthetic trajectories suitable for agent training~\cite{xu2024agenttrek, zhang2025tongui}. A distinct method empowers agents to generate trajectories by autonomously interacting with the GUI environment. This ranges from simple random or seeded exploration~\cite{zhang2025appagent, wen2024autodroid} to more sophisticated techniques like "reverse task synthesis," which infers meaningful high-level tasks from observed low-level interaction traces~\cite{sun2025genesis}.

Moving beyond using external content solely for training, another paradigm treats these resources as dynamic, non-parametric sources of knowledge accessible at runtime. This Retrieval-Augmented Generation (RAG) approach aims to enhance agent performance without necessitating model fine-tuning. For example, some work employs lightweight models to retrieve relevant guides from a knowledge base, which are then summarized to instruct a frozen agent~\cite{xu2025retrieval}. Other research focuses on mining "transition-aware knowledge" from autonomous exploration traces and formatting this knowledge into goal-oriented prompts to guide the agent's decision-making during execution~\cite{xie2025gui}.

\section{EchoTrail-GUI}
\label{sec:gui}

To address the challenges of \textit{experience acquisition} and \textit{knowledge application} in GUI agents, we introduce EchoTrail-GUI. Our framework establishes a virtuous framework of autonomous exploration, structured memory consolidation, and experience-guided inference. It operates in three core stages, as illustrated in Figure~\ref{fig:framework_overview}: (I) Critic-Guided Self-Exploration to build a memory base, (II) Dynamic Memory Injection for task-relevant retrieval, and (III) Memory-Augmented Inference for execution.

\subsection{Problem Formulation}
We formulate GUI task automation as a Partially Observable Markov Decision Process (POMDP). At each step \(t\), the agent observes the current state \(s_t\) (a GUI screenshot) and must select an action \(a_t\) from a discrete action space \(\mathcal{A}\) (e.g., click, type). Given a task instruction \(I\), the objective is to learn a policy \(\pi(a_t | s_t, I)\) that generates a trajectory of state-action pairs \(\tau = (s_0, a_0, s_1, a_1, \dots)\) to fulfill \(I\).

Standard agents rely on a frozen, pre-trained policy \(\pi_{\text{base}}(a_t | s_t, I, H_t)\), where \(H_t\) is the action history. We propose to augment this with a memory component. Our framework first constructs a memory database of successful trajectories, \(D_{\text{mem}}\). At inference time, given a new instruction \(I\), we retrieve a relevant memory set \(M_t \subset D_{\text{mem}}\). The agent's action is then sampled from a memory-augmented policy \(\pi_{\text{aug}}\):
\begin{equation}
  a_t \sim \pi_{\text{aug}}(a_t | s_t, I, H_t, M_t).
  \label{eq:augmented_policy}
\end{equation}
The central challenge, which our framework addresses, is the automated construction of a high-quality \(D_{\text{mem}}\) and the effective utilization of \(M_t\) to guide \(\pi_{\text{aug}}\).

\subsection{Stage I: Critic-Guided Self-Exploration}
\label{sec:exploration_stage}
The first stage autonomously populates the memory database \(D_{\text{mem}}\), a process detailed in Algorithm~\ref{alg:exploration}. This stage leverages an exploration agent, a critic model, and a dual-database system to generate and filter high-quality experiences without human intervention.

\subsubsection{Autonomous Trajectory Generation}

An exploration agent, \(\pi_{\text{explore}}\), built upon a foundation VLM, is responsible for generating trajectories. At each step \(t\), it takes as input the current screenshot \(s_t\), its action history \(H_t\), and real-time guidance \(G_t\) retrieved from a temporary database (see Sec~\ref{sec:dual_db}). It then outputs an action \(a_t = \pi_{\text{explore}}(s_t, H_t, G_t)\). 

To balance breadth of discovery with depth of purpose, we introduce a \textbf{Progressive Intent Focus} mechanism. The agent initiates exploration in a \textit{Curiosity-Driven Mode}, prioritizing interaction with novel UI elements. After a few steps (\(t > T_{\text{focus}}\)), it transitions to a \textit{Target-Focused Mode}, where it first formulates a concrete sub-goal (e.g., "add item to cart") and then executes actions to achieve it. This ensures that generated trajectories are both diverse and coherent.

\subsubsection{Trajectory Quality Assessment via Critic}
Upon termination, each generated trajectory \(\tau\) is evaluated by a critic, implemented as a reward model \(R_{\text{critic}}\). The critic acts as a quality filter, assigning a score based on the trajectory's coherence, efficiency, and success in achieving its implicit goal. This score is mapped to a 5-level scale. We define a quality threshold \(\theta_{\text{good}} = 4\). A trajectory is deemed high-quality and worthy of inclusion in the permanent memory if its score satisfies this condition:
\begin{equation}
    R_{\text{critic}}(\tau) \geq \theta_{\text{good}}.
    \label{eq:critic_score}
\end{equation}

Importantly, the stored trajectories are not raw screenshot sequences. Instead, to enhance generality, reduce redundancy, and maintain a lightweight memory footprint, each trajectory is converted into an abstracted representation that contains (i) a textual description of the screenshot at each step, (ii) the exploration agent’s intent summary, and (iii) the executed action. This structured, modality-agnostic format preserves the essential semantic and operational information needed for downstream retrieval and in-context guidance, while avoiding the storage cost and device-specific bias inherent in pixel-level data. In practice, an average 5-step trajectory occupies only $\sim$1k tokens in this format, compared to $>$10k tokens for an equivalent concatenated screenshot history—a $\sim$90\% reduction that makes context overhead negligible for modern VLMs with 128k context windows.

This critical filtering step is paramount, as our ablations (Sec~\ref{sec:ablation}) show that injecting low-quality memories is more detrimental than providing no memory at all.


\subsubsection{Dual-Memory Learning System}
\label{sec:dual_db}
The exploration process is scaffolded by a dual-database system to enable efficient learning:
\begin{itemize}
    \item \textbf{Processing Database (\(D_{\text{proc}}\)):} A short-term, volatile memory storing both successful and failed \textit{in-progress} trajectories. Its purpose is to provide immediate, real-time feedback (\(G_t\)) to \(\pi_{\text{explore}}\), helping it avoid repeating recent mistakes and reinforcing promising short-term strategies.
    \item \textbf{Memory Database (\(D_{\text{mem}}\)):} The final, curated, and permanent knowledge base. It contains only complete, high-quality trajectories that have passed the critic's filter. This database is the primary asset for guiding task execution in the subsequent stages.
\end{itemize}

\begin{table*}[t!]
\centering
\caption{Performance comparison on AndroidWorld benchmark. The table categorizes agents into Closed and Open Source, and indicates whether they operate without specific fine-tuning on GUI data (i.e., are 'Training-Free'). The best performance values are highlighted in \textbf{bold}. Our method is highlighted in gray.}
\label{tab:androidworld_combined_sr}

\begin{tabular}{lllcr}
\toprule
\textbf{Category} & \textbf{Agent} & \textbf{Model} & \textbf{Training-Free} & \textbf{SR$\uparrow$} \\
\midrule
\multirow{14}{*}{\begin{tabular}[c]{@{}c@{}}Closed Source\end{tabular}} 
& AppAgent~\cite{zhang2025appagent} & GPT-4o & \checkmark & 14.9 \\
& Gemini~\cite{team2024gemini} & Gemini-1.5-Pro & \checkmark & 22.8 \\
& Claude~\cite{anthropic2024Claude} & Claude Computer-Use & \checkmark & 27.9 \\
& UGround~\cite{gou2024navigating} & GPT-4o + UGround & \ding{55} & 32.8 \\
& GPT-4o~\cite{achiam2023gpt} & GPT-4o & \checkmark & 34.5 \\
& Ponder\&Press~\cite{wang2025ponder} & GPT-4o + Ponder\&Press & \ding{55} & 34.5 \\
& Aguvis~\cite{xu2024aguvis} & GPT-4o + Aguvis & \ding{55} & 37.1 \\
& M3A~\cite{rawles2024androidworld} & GPT-4o & \checkmark & 40.5 \\
& ScaleTrack~\cite{huang2025scaletrack} & GPT-4o + ScaleTrack & \ding{55} & 44.0 \\
& Aria-UI~\cite{yang2025aria} & GPT-4o + Aria-UI & \ding{55} & 44.8 \\
& URST~\cite{chenenhancingurst} & GPT-4o + URST Reflexion & \ding{55} & 46.6 \\
& AndroidGen~\cite{lai2025androidgen} & GPT-4o & \checkmark & 46.8 \\
& GUI-explorer~\cite{xie2025gui} & GPT-4o & \checkmark & 47.4 \\
& \cellcolor{gray!25}\textbf{EchoTrail-GUI} & \cellcolor{gray!25}GPT-4o & \cellcolor{gray!25}\checkmark & \cellcolor{gray!25}\textbf{51.7} \\
\midrule
\multirow{5}{*}{\begin{tabular}[c]{@{}c@{}}Open Source\end{tabular}} 
& Aguvis~\cite{xu2024aguvis} & Aguvis-72B & \ding{55} & 26.1 \\
& Qwen2.5-VL~\cite{bai2025qwen2} & Qwen2.5-VL-72B-Instruct & \checkmark & 35.0 \\
& RAG-GUI~\cite{xu2025retrieval} & RAG-GUI-72B-RSF & \ding{55} & 45.7 \\
& UI-TARS~\cite{qin2025ui} & UI-TARS-72B-SFT & \ding{55} & 46.6 \\
& \cellcolor{gray!25}\textbf{EchoTrail-GUI} & \cellcolor{gray!25}Qwen2.5-VL-72B-Instruct & \cellcolor{gray!25}\checkmark & \cellcolor{gray!25}\textbf{46.6} \\
\bottomrule
\end{tabular}
\end{table*}

\subsection{Stage II: Dynamic Memory Injection}
\label{sec:injection_stage}
This stage bridges the gap between past experience and current tasks. Given a new user instruction \(I\), we retrieve the \(K\) most relevant trajectories, \(M_t = \{\tau_1, \dots, \tau_K\}\), from \(D_{\text{mem}}\).

\subsubsection{Hybrid Retrieval Strategy}
To balance semantic understanding with keyword relevance, we employ a hybrid retrieval mechanism. We compute a relevance score for each trajectory \(\tau \in D_{\text{mem}}\) as a weighted combination of dense and sparse retrieval scores:
\begin{equation}
\text{Score}(\tau, I) = \alpha \cdot S_{\text{dense}}(\tau, I) + (1-\alpha) \cdot S_{\text{sparse}}(\tau, I),
\end{equation}
where \(S_{\text{dense}}\) is the cosine similarity between the embeddings of \(I\) and the final intent of \(\tau\) (computed via FAISS), and \(S_{\text{sparse}}\) is the BM25 lexical-match score. We retrieve the top-\(K\) trajectories based on this score. Our experiments show \(K=2\) yields an optimal trade-off (see Sec~\ref{sec:k_sensitivity_analysis}).

\subsubsection{Memory Formatting}
Raw trajectories are not directly consumable. Each retrieved trajectory \(\tau_k \in M_t\) is processed into a structured, human-readable format. This involves summarizing each step into a tuple of \textit{\{interface description, agent intent, action\}}. This transformation converts a raw log into an intuitive, step-by-step guide, making it an effective in-context learning example.

\subsection{Stage III: Memory-Augmented Inference}
\label{sec:inference_stage}
In the final stage, the retrieved and formatted memories are used to guide a GUI agent, \(\pi_{\text{agent}}\), which can be any off-the-shelf VLM. This highlights the plug-and-play nature of our framework.

At each inference step \(t\), the agent's behavior is conditioned on a structured prompt \(P_t\), which integrates multiple sources of information:
\begin{equation}
    P_t = f(I, M_t, H_t, s_t, E_{\text{sum}}(s_t)),
\end{equation}
where \(f\) is a template function, \(I\) is the task instruction, \(M_t\) are the formatted memories, \(H_t\) is the current action history, \(s_t\) is the screenshot, and \(E_{\text{sum}}\) is a summarization model that provides a textual description of \(s_t\). The agent then produces the next action \(a_t \sim \pi_{\text{agent}}(P_t)\). By providing explicit, successful examples (\(M_t\)), we effectively guide the agent's reasoning process. The agent itself determines task completion by outputting a special `finish` action from its augmented action space \(\mathcal{A} \cup \{\text{finish}\}\).

\section{Experiments}
\label{sec:exp_new}
In this section, we present a rigorous empirical evaluation of the EchoTrail-GUI framework. Our experiments are designed to answer three primary research questions: (1) Does EchoTrail-GUI outperform state-of-the-art GUI agents on established benchmarks? (2) What is the quantifiable contribution of each component within our framework? (3) How effective and efficient is our proposed self-exploration mechanism?

\subsection{Experimental Settings}

\subsubsection{Environments and Baselines}
We conduct our evaluation on two widely-adopted, interactive Android benchmarks: \textbf{AndroidWorld}~\cite{rawles2024androidworld}, which features a diverse set of real-world applications, and \textbf{AndroidLab}~\cite{xu2024androidlab}, providing a more controlled environment for systematic testing. 

Our framework's performance is compared against a comprehensive suite of recent and state-of-the-art agents. This includes powerful proprietary models like GPT-4o and Gemini-1.5-Pro, as well as specialized open-source agents such as UGround~\cite{gou2024navigating} and UI-TARS~\cite{qin2025ui}. To demonstrate the model-agnostic nature of EchoTrail-GUI, we implement our framework on two distinct backbones: a leading open-source model, Qwen2.5-VL-72B-Instruct, and a leading proprietary model, GPT-4o.

\subsubsection{Evaluation Metrics}
\label{sec:metrics}

\begin{table*}[t!]
\centering
\caption{Performance comparison on AndroidLab benchmark. The table categorizes agents into Closed and Open Source, and indicates whether they operate without specific fine-tuning on GUI data (i.e., are 'Training-Free'). The best performance values are highlighted in \textbf{bold}. Our method is highlighted in gray.}
\label{tab:androidlab_performance_combined}
\resizebox{\textwidth}{!}{%
\begin{tabular}{lllcccccc}
\toprule
\textbf{Category} & \textbf{Agent} & \textbf{Model} & \textbf{Training-Free} & \textbf{Sub-SR} & \textbf{RRR} & \textbf{ROR} & \textbf{SR$\uparrow$} \\
\midrule
\multirow{6}{*}{Closed Source} 
& Gemini~\cite{team2023gemini} & Gemini-1.0 & \checkmark & 12.6 & 72.5 & 76.7 & 10.9 \\
& Claude~\cite{anthropic2024Claude} & Claude-3-Opus & \checkmark & 15.1 & 81.4 & 83.9 & 13.0 \\
& Gemini~\cite{team2024gemini} & Gemini-1.5-Pro & \checkmark & 18.5 & 106.0 & \textbf{91.5} & 16.7 \\
& Claude~\cite{anthropic2024Claude} & Claude-3.5-Sonnet & \checkmark & 32.7 & \textbf{113.4} & 81.2 & 29.0 \\
& GPT-4o~\cite{achiam2023gpt} & GPT-4o & \checkmark & 35.0 & 87.3 & 85.4 & 31.2 \\
& AutoGLM~\cite{liu2024autoglm} & AutoGLM-2024-10 & \ding{55} & --- & --- & --- & 36.2 \\
\rowcolor{gray!25} 
& \textbf{EchoTrail-GUI}  & GPT-4o & \checkmark & \textbf{50.7} & 97.9 & 88.5 & \textbf{48.1} \\

\midrule
\multirow{6}{*}{Open Source} 
& UI-TARS~\cite{qin2025ui} & UI-TARS-72B & \ding{55} & 12.0 & 40.8 & 70.0 & 9.6 \\
& UI-TARS~\cite{qin2025ui} & UI-TARS-72B-ft & \ding{55} & 28.4 & 81.4 & 81.6 & 22.1 \\
& Qwen2.5-VL~\cite{bai2025qwen2} & Qwen2.5-VL-72B-Instruct & \checkmark & 26.1 & 68.7 & 81.4 & 23.9 \\
& Qwen2-VL~\cite{wang2024qwen2} & Qwen2-VL-72B-Instruct-ft & \ding{55} & 29.3 & 84.5 & 90.2 & 24.6 \\
& Qwen2.5-VL~\cite{bai2025qwen2} & Qwen2.5-VL-72B-Instruct-ft & \ding{55} & 30.9 & 81.3 & 79.3 & 25.0 \\
\rowcolor{gray!25} 
& \textbf{EchoTrail-GUI}  & Qwen2.5-VL-72B-Instruct & \checkmark & \textbf{41.1} & \textbf{89.4} & \textbf{92.1} & \textbf{37.5} \\
\bottomrule
\end{tabular}%
}
\end{table*}

To ensure a comprehensive assessment, we adopt a multi-faceted set of metrics from the AndroidLab benchmark. SR, Sub-SR, RRR, and ROR stand for Success Rate, Sub-Goal Success Rate, Reversed Redundancy Ratio, and Reasonable Operation Ratio, respectively.

\begin{itemize}
    \item \textbf{SR\%:}Success Rate. The primary metric, representing the percentage of tasks completed successfully.
    \item \textbf{Sub-SR\%:}Sub-Goal Success Rate. The percentage of correctly completed sub-goals, which measures fine-grained progress within a task.
    \item \textbf{RRR\%:}Reversed Redundancy Ratio. Measures the agent's step efficiency. A higher RRR indicates that the agent took fewer redundant steps compared to an expert trajectory.
    \item \textbf{ROR\%:}Reasonable Operation Ratio. Measures the quality of individual actions by calculating the proportion of operations that are deemed ``reasonable'' or non-erroneous, indicating operational robustness.
\end{itemize}
For the AndroidWorld benchmark, we report the standard Success Rate.

\subsubsection{Implementation Details}
\label{sec:implementation}

Our framework's components are implemented with the following specifications:
\begin{itemize}
    \item \textbf{Exploration Agent:} Powered by Gemini 2.5 Flash. Maximum trajectory length is 30 steps.
    \item \textbf{Reward Model:} Implemented with Gemini 2.5 Flash Lite, assigning a 1-5 score where the quality threshold \(\theta_{\text{good}}\) is 4.
    \item \textbf{Inference Agent:} We use Qwen2.5-VL-72B-Instruct and GPT-4o as backbones.
    \item \textbf{Summarization \& Embedding:} We utilize Qwen3-30B-Instruct-2507 for summaries and Qwen3-Embedding-4B for embeddings.
    \item \textbf{Retrieval:} A hybrid FAISS and BM25 system retrieves \(K=2\) memories, a choice justified by our sensitivity analysis in Section~\ref{sec:k_sensitivity_analysis}.
    \item \textbf{Exploration Scale:} The EchoTrail-4K dataset was constructed over 4{,}143 episodes with an average trajectory length of 4.8 steps (maximum 22 steps), requiring approximately 4.3 minutes per valid trajectory on a single device. Retrieval via FAISS adds negligible latency during inference.
\end{itemize}

\subsection{Main Results}
We first evaluate the end-to-end performance of EchoTrail-GUI against all baselines. As a training-free, plug-and-play framework, our approach provides a substantial performance uplift to existing models by simply augmenting them with experiential memory. We note that recent leaderboard entries (e.g., DroidRun, AGI-0) often rely on extensive pre-training, fine-tuning, or privileged APIs that differ from the standard AndroidWorld evaluation protocol; accordingly, our results and claims are scoped to \textbf{training-free agents evaluated under the standard setting}. The detailed results are presented in Table~\ref{tab:androidworld_combined_sr} and Table~\ref{tab:androidlab_performance_combined}.

\subsubsection{Performance on AndroidWorld}
As shown in Table~\ref{tab:androidworld_combined_sr}, EchoTrail-GUI establishes itself as a leading framework on AndroidWorld. When applied to GPT-4o, our framework yields substantial performance gains (51.7\% SR), significantly outperforming the base model and surpassing other augmentation methods that require task-specific training. This strong trend holds in the open-source category: EchoTrail-GUI elevates Qwen2.5-VL-72B-Instruct to match top fine-tuned agents (46.6\% SR) without any model retraining, validating the power of our memory-driven approach. The gains are particularly pronounced on medium- and hard-difficulty tasks, which typically require multi-step reasoning over unfamiliar application workflows—precisely the regime where retrieved experience provides the most informative prior.

\subsubsection{Performance on AndroidLab}
The results on AndroidLab (Table~\ref{tab:androidlab_performance_combined}) corroborate these findings across a more comprehensive set of metrics. On Qwen2.5-VL, EchoTrail-GUI more than doubles the SR and delivers consistent improvements across all dimensions—sub-goal completion, step efficiency, and operational robustness—indicating the agent becomes both more successful and more direct in its actions. A similar pattern holds for GPT-4o, confirming the generality of our approach across diverse tasks and environments.

\subsubsection{Cross-Benchmark Discussion}
Across both benchmarks, a consistent pattern emerges: EchoTrail-GUI provides the largest gains on complex, multi-step tasks where prior experience is most valuable. On AndroidLab, the improvement in Sub-SR (+15.0 on Qwen2.5-VL) indicates that memory injection helps the agent complete more intermediate sub-goals, not just final objectives. The simultaneous improvement in RRR and ROR metrics further suggests that retrieved trajectories reduce both redundant actions and erroneous operations, leading to more efficient and reliable task execution overall. Importantly, these gains are achieved without any model retraining, highlighting the practical value of non-parametric memory as a lightweight yet effective augmentation strategy.

\subsection{Ablation and In-Depth Analysis}
Having established the overall effectiveness of EchoTrail-GUI, we now dissect the framework to understand the contribution of its individual components and the dynamics of the self-exploration process.

\subsubsection{Ablation Studies}
\label{sec:ablation}
We perform ablation studies on AndroidWorld using Qwen2.5-VL-72B-Instruct (Table~\ref{tab:ablation_studies}). The full system provides a substantial improvement over the backbone model. Among all factors, removing \textit{Critic-based Filtering} causes the largest drop—performance falls \emph{below} the no-memory baseline, confirming that unfiltered, low-quality trajectories actively mislead the agent and are more harmful than omitting memory entirely.

Removing \textit{Hybrid Retrieval} also degrades results, as random or poorly matched examples fail to provide meaningful guidance and may introduce conflicting signals. Disabling \textit{Real-time Guidance} weakens exploration quality, yielding a less diverse memory base and consequently less effective downstream inference. Each component thus plays a complementary role: critic filtering ensures reliability, hybrid retrieval ensures relevance, and real-time guidance ensures quality and coverage.

\begin{table}[t!]
\centering
\caption{Ablation study on AndroidWorld. We start with the backbone model and show the performance of our full framework, followed by variants with specific components removed.}
\label{tab:ablation_studies}
\footnotesize
\setlength{\tabcolsep}{4pt}
\resizebox{\columnwidth}{!}{%

\begin{tabular}{p{3.3cm}cccc} 
\toprule
\multirow{2}{*}{\textbf{Method}} & \multicolumn{4}{c}{\textbf{AndroidWorld}} \\
\cmidrule(lr){2-5}
& \textbf{Easy} & \textbf{Medium} & \textbf{Hard} & \textbf{Avg. SR (\%)} \\
\midrule
Qwen2.5-VL-72B-Instruct & 46.7 & 23.6 & 13.2 & 34.1 \\
\quad \textit{w/o Critic-based Filtering} & 47.5 & 13.9 & 10.5 & 31.0 \\
\quad \textit{w/o Hybrid Retrieval} & 60.7 & 20.8 & 13.2 & 40.5 \\
\quad \textit{w/o Real-time Guidance} & 62.3 & 25.0 & 13.2 & 42.7 \\
\rowcolor{gray!25}
\textbf{EchoTrail-GUI} & \textbf{65.6} & \textbf{30.6} & \textbf{15.8} & \textbf{46.6} \\
\bottomrule
\end{tabular}%
}

\end{table}

\begin{figure}[t!]
    \centering
    \includegraphics[width=\columnwidth]{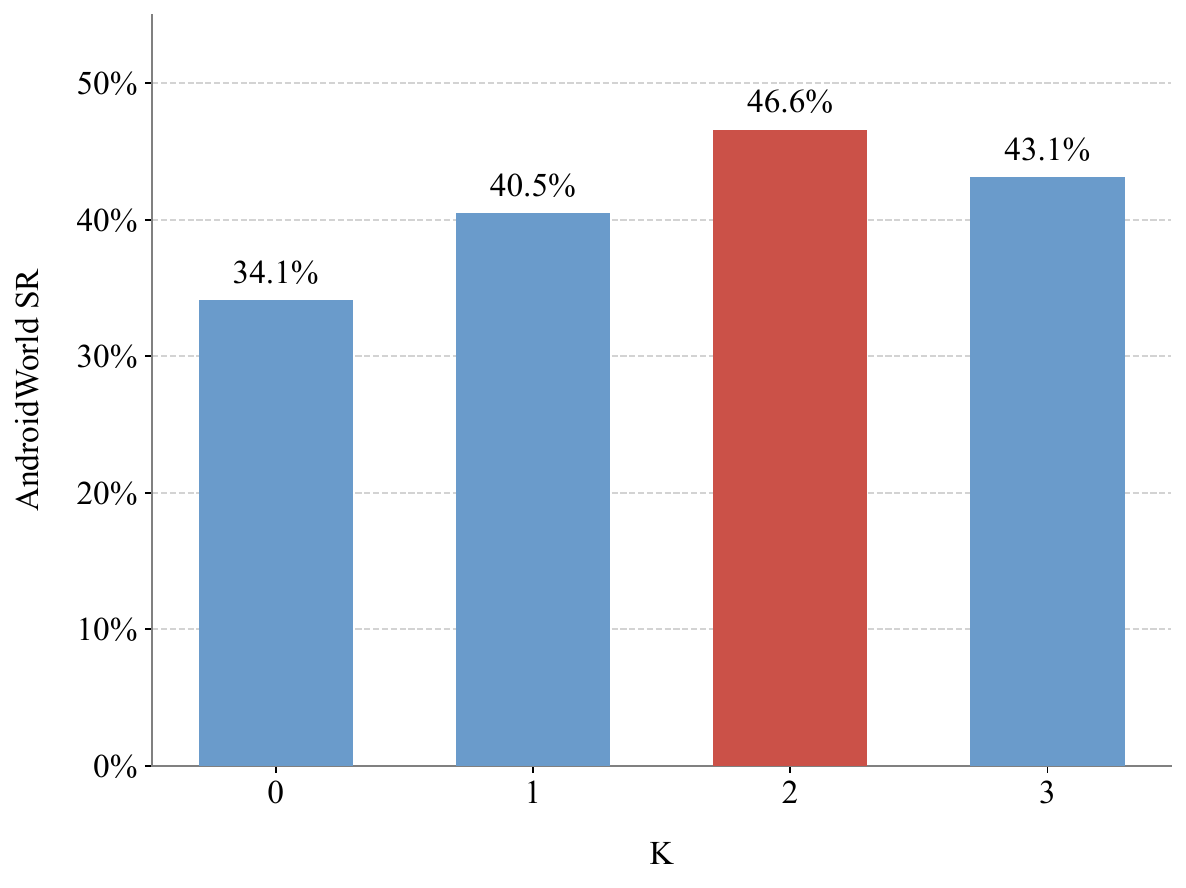}
    \caption{Sensitivity analysis of the number of injected memories (\(K\)) on AndroidWorld success rate.}
    \label{fig:k_sensitivity}
\end{figure}

\begin{figure}[t!]
    \centering
    \includegraphics[width=\columnwidth]{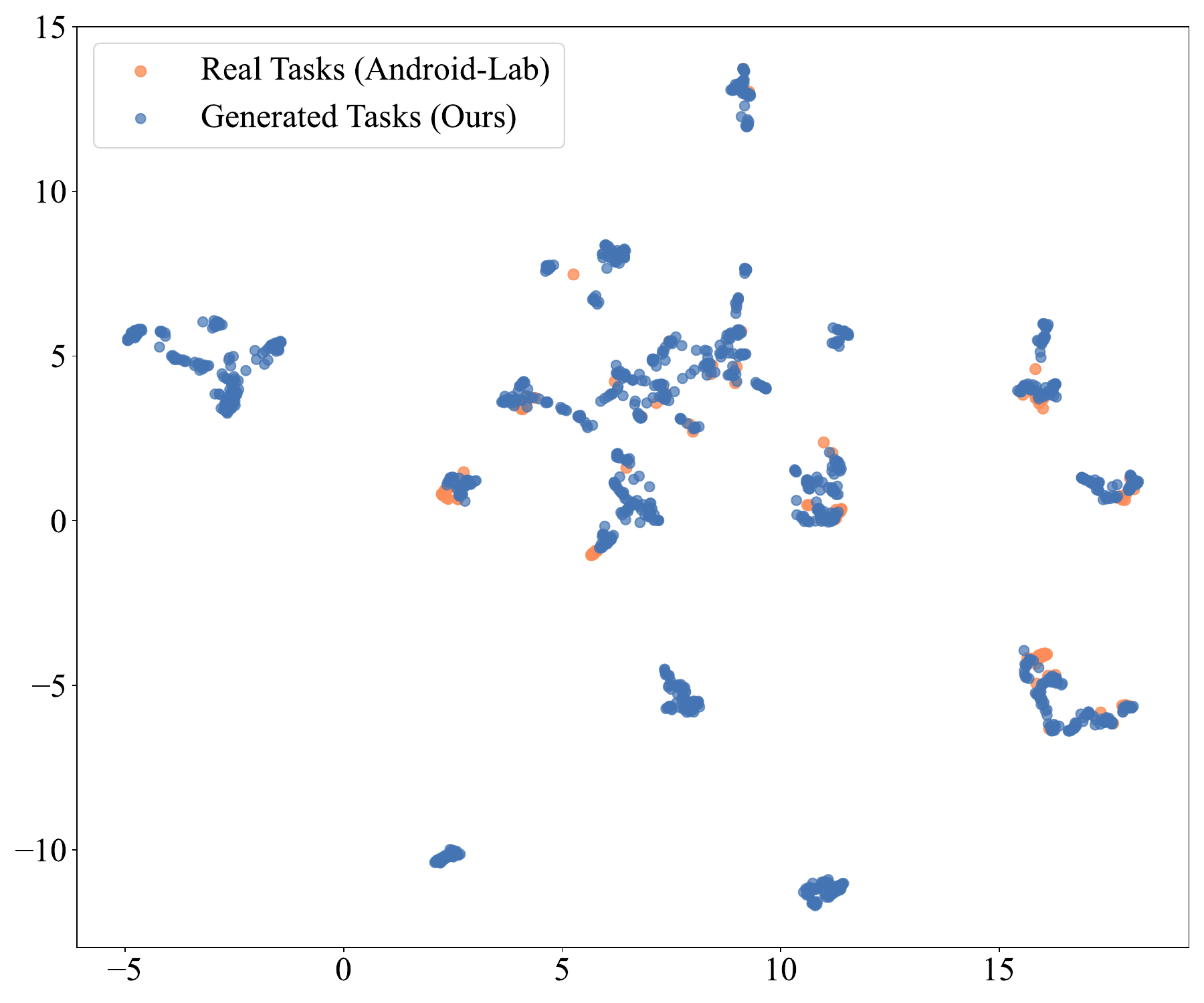}
    \caption{UMAP visualization of task instructions from the AndroidLab test set (orange) and our self-explored trajectories (blue), demonstrating high semantic alignment and diversity.}
    \label{fig:umap_visualization}
\end{figure}

\subsubsection{Analysis of the Self-Exploration Mechanism}
A core novelty of our work is the autonomous construction of the memory base. We analyze two key aspects of this process.

\paragraph{Quality and Diversity of Generated Trajectories.}
We validate that our exploration produces realistic and diverse trajectories by encoding the final task intents from EchoTrail-4K with Sentence-BERT~\cite{reimers2019sentence} and comparing them against AndroidLab ground-truth instructions via UMAP~\cite{mcinnes2018umap} (Figure~\ref{fig:umap_visualization}). The dense overlap between generated (blue) and real (orange) tasks confirms \textit{realism}: the agent generates tasks that are structurally and intentionally similar to human-defined goals, rather than random or nonsensical sequences. Meanwhile, the generated task space extends significantly beyond the boundaries of the original test set, filling in gaps and exploring new semantic regions. This confirms the \textit{diversity} and effectiveness of our exploration strategy in building a comprehensive experience base that enhances generalization.

\begin{figure}[h!]
    \centering
    \includegraphics[width=\columnwidth]{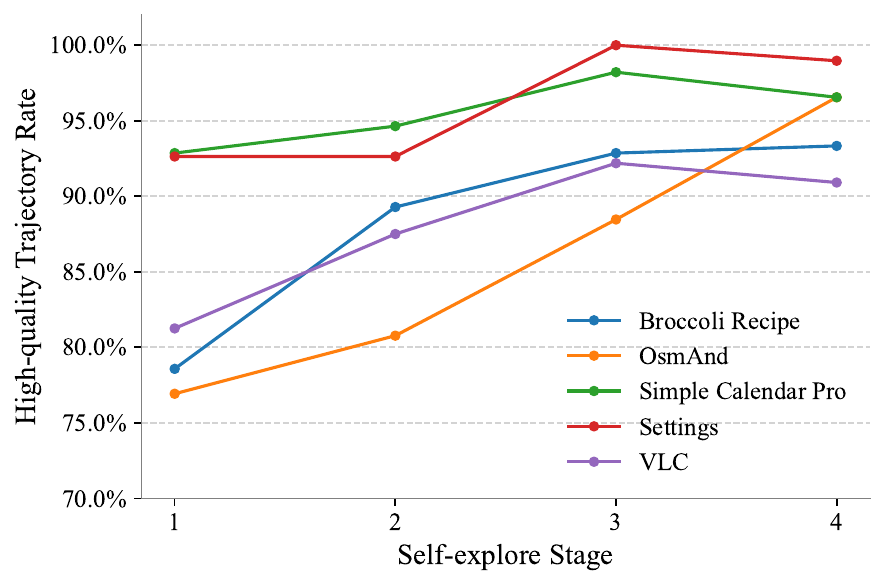}
    \caption{The rate of high-quality trajectories generated across four sequential exploration stages on representative apps from AndroidWorld.}
    \label{fig:quality_over_time}
\end{figure}

\paragraph{Effectiveness of the Real-time Guidance Mechanism.}
\label{sec:quality_over_time}
We track the proportion of high-quality trajectories across four sequential exploration stages for representative apps (Figure~\ref{fig:quality_over_time}). A clear upward trend is observed across all applications: complex apps like ``OsmAnd'' and ``VLC'' show $\sim$20 percentage-point gains, demonstrating rapid adaptation to unfamiliar interfaces. For more intuitive applications like ``Settings'', the framework consolidates already-high performance. This self-correcting behavior confirms the efficacy of our dual-database learning system in progressively improving exploration quality and populating the memory database with reliable experiences.

\subsubsection{Analysis of Memory Injection}
\label{sec:k_sensitivity_analysis}
We analyze the effect of the number of retrieved memories $K$ on performance (Figure~\ref{fig:k_sensitivity}). Injecting a small number of relevant trajectories consistently improves over the memory-free baseline ($K=0$), confirming the value of targeted experiential context. Performance peaks at $K=2$; beyond this point, additional trajectories yield diminishing returns and eventually degrade performance due to increased prompt length, conflicting advice, and contextual dilution. We therefore select $K=2$ for all main experiments as a practical compromise between informativeness and efficiency.

\subsubsection{Critic Reliability Analysis}
\label{sec:critic_reliability}
A key design assumption of EchoTrail-GUI is that the critic can serve as a reliable, scalable proxy for human judgment. We validate this along two dimensions.

\paragraph{Human Alignment.}
We compare critic decisions against human expert labels on 100 stratified trajectories. The two exhibit Cohen's $\kappa \approx 0.72$, indicating \emph{substantial agreement} (the standard threshold is $\kappa \geq 0.61$). This confirms that evaluating abstracted Intent/Action sequences via textual reasoning—rather than raw pixels—is sufficiently reliable for high-quality memory filtering.

\paragraph{Model Agnosticism.}
To demonstrate that proprietary APIs are not mandatory, we evaluate four alternative models as Critics against our default (Gemini). As shown in Table~\ref{tab:model_consistency}, the open-source \textbf{Qwen3-VL-235B-A22B} achieves $\text{F1}=0.86$ and $\kappa=0.68$, comparable to GPT-4o, confirming that open-weight models can reliably replace proprietary critics without structural changes.

\begin{table}[h!]
\centering
\caption{Consistency of alternative Critic models against the default Gemini baseline ($N=100$).}
\label{tab:model_consistency}
\resizebox{7cm}{!}{%
\footnotesize
\setlength{\tabcolsep}{4pt}
\begin{tabular}{lcccc}
\toprule
\textbf{Model} & \textbf{Acc.} & \textbf{$\kappa$} & \textbf{Prec.} & \textbf{F1} \\
\midrule
Qwen3-VL-235B-A22B & 0.84 & 0.68 & 0.77 & 0.86 \\
GPT-4o mini       & 0.83 & 0.66 & 0.79 & 0.84 \\
GPT-4o            & 0.81 & 0.62 & 0.77 & 0.82 \\
Qwen3-VL-30B-A3B   & 0.80 & 0.60 & 0.74 & 0.82 \\
\bottomrule
\end{tabular}%
}
\end{table}

\section{Conclusion}
\label{sec:conclusion}

We introduced EchoTrail-GUI, a unified framework that addresses the ``digital amnesia’’ of contemporary GUI agents by enabling them to autonomously construct and leverage actionable memory. Through critic-guided self-exploration, lightweight trajectory abstraction, and dynamic memory injection, the framework establishes an effective \emph{learn–remember–apply} cycle that transforms stateless VLM-based agents into adaptive, experience-aware systems. Experiments on AndroidWorld and AndroidLab demonstrate substantial gains in success rate, sub-goal completion, and operational efficiency, while ablations confirm the critical role of each component. Our results show that reliable operational memory can be built without manual supervision and that abstracted representations—comprising interface descriptions, inferred intentions, and executed actions—provide a scalable and transferable basis for contextual reasoning, positioning EchoTrail-GUI as a practical path toward GUI agents that continually improve through accumulated experience. Beyond the specific benchmarks studied, we believe the core principle—that autonomous self-exploration guided by a lightweight critic can produce high-quality, reusable experience—is broadly applicable and could benefit other interactive AI systems facing similar statelessness challenges.


{
    \small
    \bibliographystyle{ieeenat_fullname}
    \bibliography{main}
}




\end{document}